# Edge Detection for Pattern Recognition: A Survey

Alex Pappachen James
School of Engineering
Nazarbayev University
E: apj@ieee.org

## Abstract

This review provides an overview of the literature on the edge detection methods for pattern recognition that inspire from the understanding of human vision. We note that edge detection is one of the most fundamental process within the low level vision and provides the basis for the higher level visual intelligence in primates. The recognition of the patterns within the images relate closely to the spatiotemporal processes of edge formations, and its implementation needs a crossdisciplanry approach in neuroscience, computing and pattern recognition. In this review, the edge detectors are grouped in as edge features, gradients and sketch models, and some example applications are provided for reference. We note a significant increase in the amount of published research in the last decade that utilizes edge features in a wide range of problems in computer vision and image understanding having a direct implication to pattern recognition with images.

## Introduction

Understanding and identifying object features from images in an accurate and computationally efficient way is one of foremost modern challenges in pattern recognition with images(Dell'acqua, Roberto, & Remo, 1998; Grauman & Leibe, 2010; Konash & Dmitri, 1989; Logothetis, 1996; Mamic & Mohammed, 2000; Nguyen, Nebel, & Florez-Revuelta, 2016; Rigelsford & Jon, 2003; Treiber & Marco, 2010; G. Xu & Zhang, 2013). Pixels in an image contain unique information about the objects that once extracted through various mathematical transforms is often referred to as object features(Akagunduz, Erdem, & Ilkay, 2011; C., Rory, Bakker, & Claire, 2011; R. S. T. Lee & Liu, 2003; F. Li, Fangxing, Linyu, & Huafei, 2011). The foundation feature that defines an image is the pixel changes defined as the spatial differences, which in a mathematical sense in an image space can be represented as region wise solutions of differential equations of different orders(Cappellini, 1997; Chou & Wen-Shou, 1999; Gauman & Leibe, 2010; "Object Recognition," n.d.; Ponce, Hebert, Schmid, & Zisserman, 2007; Rosenfeld, Azriel, Saha, & Akira, 2001; Si & Zhu, 2013).

The primary ideas on these pixel difference calculations originate from the neural studies of human vision processing systems (Bundy, Alan, & Lincoln, 1984; Cheng-en Guo, Cheng-en, Song-Chun, & Wu, 2003; Duperthuy, Christophe, & Jean-Michel, 1997; Lehky & Tanaka, 2016; Lindeberg & Tony, 1994; David Marr & Hildreth, 1979). The early understanding of the vision processing system on primates considers the sensory neural processing mechanisms of the human eye, and how the signals in the eye are generated for a given set of stimulus(Burr, Concetta Morrone, & Donatella, 1989; Erlikhman & Kellman, 2015; Hood & Birch, 1993; Meese & Freeman, 1995; Shapley & Tolhurst, 1973; Tolhurst, 1972; Watt & Morgan, 1983). There are strong evidences that the edge detection is an important phenomenon in human visual system that helps to define the objects and therefore its recognition(Burr et al., 1989; Georgeson, May, Freeman, & Hesse, 2007; Meese & Freeman, 1995; Watt & Morgan, 1983).

The recognition of the patterns offer several practical challenges of detection of edges in the digital images. In practise the digital pixel is square in shape, and its pixel intensity accuracy is restricted by the sensor hardware, digital memory and sampling rates(D. G. Chen, Law, Lian, & Bermak, 2014; Yong Chen, Yong, Fei, & Gul, 2009; Chia-Nan Yeh, Chia-Nan, Yen-Tai, & Jui-Yu, 2008; Fesenmaier & Catrysse, 2008; Gottardi, Sartori, & Simoni, 1993; Ho, Derek, Glenn, & Roman, 2012; Jayaraman, 2011; M.-K. Kim, Hong, & Kwon, 2015; Kleinfelder, Suki-Iwan, Xinqiao, & El Gamal, n.d.; Sarkar & Theuwissen, 2012; Shi, Zhouyuan, & Martin, 2006; Skorka, Orit, & Dileepan, 2014; C. Zhang, Chi, Suying, & Jiangtao, 2011; Zhao et al., 2014). These restrictions often leads to information loss within and between the image pixels, and that along the object edges. The continuous edges of the objects no longer looks accurate in space, reflected as missing information in discrete spatial changes. These basic problems in digital image formation process adopted in modern implementations of imaging camera and computers makes the edge detection one of the most scientifically difficult problem to solve. And since edges form the basic sources of features in the objects for a wide range of pattern recognition tasks such as face recognition(Datta, Datta, & Banerjee, 2015; Alex P. James, 2013; Alex Pappachen James & Sima, 2008; Maheshkar, Vikas, Suneeta, Srivastava, & Sushila, 2012; Rodrigues & du Buf, 2009; Yang, Fan, Michel, & Hervé, 1996), iris recognition(Bodade & Talbar, 2014; Burge & Bowyer, 2013; Dori, Dov, & Haralick, 1995; Robinson, n.d.; Wildes & Richard, n.d.), fingerprint recognition(Al-Dulaimi, 2013; Jain, Halici, Hayashi, Lee, & Tsutsui, 1999; Kulshrestha, Megha, Banga, & Sanjeev, 2012; Maltoni, Maio, Jain, & Prabhakar, 2009; Mishra, Parul, Shrivastava, & Amit, 2013; Ratha & Bolle, 2003), activity detection, object detection and tracking(Chartier & Lepage, n.d.; Dollar, Zhuowen, & Belongie, n.d.; Dongcheng Shi, Dongcheng, Liqang, & Ying, n.d.; Gruen, Armin, & Dirk, 1993; Krammer & Schweinzer, 2006; Mae & Shirai, n.d.; Rodrigues & du Buf, 2009; Shirai, Yoshiaki, Yasushi, & Shin'ya, 1996), this becomes one of the most important topic in applied pattern recognition.

# Edge Models

There exists various inter-related theories on edge formation in human eye and subsequent computational models feasible for implementation in a digital computer. The models can be grouped into sketch models, and its deduced forms of edge features and its implementation techniques from gradient filters. Figure 1 shows the grouping of the methods based on the understanding of image edges in the perspective of neuroscience, pattern analysis and signal processing.

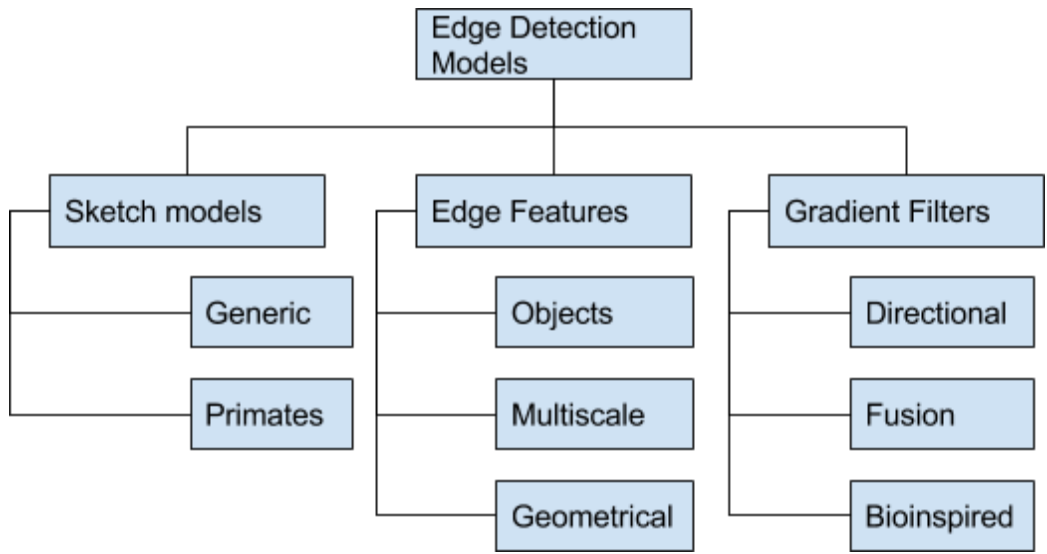

Figure 1: A grouping of edge detection based on the theoretical understanding with respect to neuroscience, pattern analysis and signal processing, that can be used for pattern recognition problems.

## Sketch models

The early edge detection schemes inspire from the sketch models (David Marr & Hildreth, 1979) that looks at a physiological model of zero crossing detection. The implementation of the early zero crossing detection methods show a mathematical description of convolution filters (Hildreth, 1980; King, 1985). An early account of edge detection techniques (Argyle & Rosenfeld, 1971) describe the sketching abilities of the human brain, and its translations to computationally feasible models. Whiles these theories are largely attempted to be understood in the neural context and is an interesting topic for vision research, they extend the applications to deep into the very basic understanding of image processing and subsequently the quality of features in pattern recognition.

While, edges are considered as important from the very early days of imaging studies, the deeper understanding of the features themselves serve as an ongoing research problem(Herman M. Gomes & Fisher, 2003; Guo, 2005; Korn, 1989; Lindeberg, 1991; Q. Li & Qi, 2014; Morgan, 2011). The computationally feasible models of the sketches have several possible applications in the pattern recognition(H. M. Gomes & Fisher, n.d.; D. Marr, 1980; Van De Ville & Unser, 2008). The theory of the sketches can be extended to other well known signal processing methods and filters to mimic the various levels of the edge processing(D. Marr, 1980; Van De Ville & Unser, 2008).

## Edge features

The inference of the edges from being a descriptor to identify the structure to being a feasible feature for pattern matching requires the understanding of human vision (Georgeson et al., 2007). The edge features of the objects has moved from being a digital only representation to a discrete edge representation in digital images, and often is highly sensitive to variations in natural variability within the images. The most common natural variability that have a significant impact on the quality of the edge features are illumination changes and sensor noises. Whereas, that would affect the recognition of the objects using the feature also depend on the scale and shift sensitivity of the edges, the pose of the objects, the quality and quantity of training image set(Chartier & Lepage, n.d.; Dollar et al., n.d.; Dongcheng Shi et al., n.d.; Gruen et al., 1993; Krammer & Schweinzer, 2006; Lin & Kung, 2000; Mae & Shirai, n.d.; Robinson, n.d.; Shirai et al., 1996).

Computationally efficient implementation of the edges is also a serious issue to consider in the real-time detection and extraction of object edges(Alex, Asari, & Alex, 2013; Dollár & Zitnick, 2015; King, 1985; Latecki, 1998; Mathew & James, 2015). The ability for the edges to become robust to noise is an important issues when the edge features is used to perform pattern recognition tasks such a face and face sketch recognition (Alex, Asari, & Alex, 2012; Alex et al., 2013; R. Hu, Rui, & John, 2013).

# Gradient Filters

The implementation of the method for extraction of the edge features and calculation of the edges in digital computers are done by applying convolution filtering techniques. The most common approach is gradient filtering with different window coefficients representing implementation of different types of the discrete pixel difference equations(Dasarathy & Dasaratky, n.d.; Douglas, n.d.-a, n.d.-b; "Edge filters," n.d., "Gradient Filters," n.d.; Mathews & Xie, n.d.; Terol‑Villalobos, 1996). The preservation of the edges in noisy images is the most challenging task under such filtering schemes(Bachy & Zaidi, 2015; Bourne, 2010; Burger & Burge, 2012; Faghih & Smith, 2002; Hornberg, 2007; Jacob & Unser, 2004; Kerre & Nachtegael, 2013; Y. Lee & Yunwoo, 2000; Nezhadarya & Ward, 2011; O'Gorman, Sammon, & Seul, 2008; Ren, Lei, Dai, & Li, 2015; Sinha & Dougherty, 1998).

The methods such as (Alex et al., 2012, 2013; R. Hu et al., 2013; Mathew & James, 2015; Van De Ville & Unser, 2008) making use of the psychological mechanism of human mind and brain tends to outperform the gradient only approaches indicating the benefit of drawing inspiration from the neuroscience studies to the benefit of image processing and pattern recognition studies. Another frequent and useful approach to generate useful features from the gradients is to combine the different filters to achieve fused features having several aspects of the edge information useful for pattern matching (George & Unnikrishnan, 2016; Ghantous, Milad, Soumik, & Magdy, 2008; Ghassemian, n.d.; J. Hu, Jianwen, & Shutao, 2012; Kwon Lee, Kwon, & Simon, 2015; Petrovic & Xydeas, 2004; Pritika, Pritika, & Sumit, 2015; Stathaki, 2011; Stienne, Reboul, Azmani, Choquel, & Benjelloun, 2014; X. Wang, 1992; Woo, 1998; X. Xu, Xin, Qiang, & Deshen, 2010; Zeng, Tao, Changyu, & Fei, 2012; Z. Zhang, Zhike, Weiqiang, & Ke, 2014; Zulkifley, Moran, & Rawlinson, 2012).

# Applications

The most common application of the edge detection in the recognition of the patterns from the images are that for segmentation (Dasarathy & Dasaratky, n.d.), region identification(Akram, Kim, Lee, & Choi, 2015; Sen, Debashis, & Pal, 2010; K. Zhang, Kedai, Hanqing, Miyi, & Qi, 2006), feature detection(Georgeson et al., 2007; Ren et al., 2015), object searching(Frintrop, 2006; Kmieć, Marcin, & Andrzej, 2015; Leeds, Daniel, & Michael, 2015; P. Wang & Ping, 2010), object tracking(Gao, Parslow, & Tan, n.d.; Wei Jyh Heng & Ngan, 2001; W. J. Heng & Ngan, n.d.; Y. Liu, Ya, Haizhou, & Guang-you, 2001; Mukherjee, Potdar, & Potdar, 2010; "Object Detection and Tracking," 2013; Shen, Pankanti, & Wang, 2001; P. Wang & Ping, 2010) on various problems in medical imaging(Avramovic & Aleksej, 2011; Dua & Sumeet, 2010; Guddanti, 1997; Leondes, 2003; "Mean Curvature Flows, Edge Detection, and Medical Image Segmentation," n.d.; Riste-Smith, 1990), biometrics(Drahansky & Martin, 2011; Pflug & Busch, 2012; J. Wang et al., 2011; D. Zhang & Jain, 2006) and object recognition(Kmieć et al., 2015; Robinson, n.d.; Rodrigues & du Buf, 2009).

# Biometric recognition

Face images contain distinct biometric information from the boundaries of eyes, nose, mouth, and jaw shape, that can be represented with image edges (Rodrigues & du Buf, 2009; Yang et al., 1996). The features inspired from the spatial change detection (Alex Pappachen James & Sima, 2008) that results in feature edges have shown high robustness to natural variabilities(Alex P. James, 2013; Pappachen James, James, & Sima, 2010). The noise that usually makes the feature extraction of the edges extremely difficult task would lead to poor performance of face recognition (Alex et al., 2012, 2013; R. Hu et al., 2013; Mathew & James, 2015). The use of psychometric measures in the calculation of edges have shown to improve the robustness to the face recognition (Mathew & James, 2015).

Optimisation techniques to enhance the edges in the images is another open issue in pattern recognition (Setayesh, Mahdi, Mengjie, & Mark, 2011). The use of fractal imaging techniques and particle swarm optimisation techniques have shown to improve the edge detection results (Demers & Matthew, 2012; Setayesh, Mahdi, Mengjie, & Mark, 2010).

The importance of a good edge detection method is of paramount importance to problems such as fingerprint recognition (L. Zhang & Liang, 2014). In a fingerprint, the features can encoded in a binary form or as in a transformed domain to extract the minutiae (Bigun & Josef, 2014; L. Zhang & Liang, 2014). There are several mathematical transforms used to extract the edge features or the coordinates of the features for its matching ("A Robust Fingerprint Matching System Using Orientation Features," 2015; Balti, Ala, Mounir, & Farhat, 2012, 2014; Yi Chen, Yi, & Jain, 2007; Chung et al., 2005; Dale & Joshi, 2008; He, Tian, Li, He, & Yang, 2007; Kumar, Ajay, & Yingbo, 2011; Tachaphetpiboon & Amornraksa, 2007, n.d.; Tang & Ting, 2012; Xie Meihua, Xie, & Wang, n.d.).

# Object recognition

The boundary of the objects in movements represented as directional edges have important cues for the recognition and tracking (Das, Dipankar, Yoshinori, & Yoshinori, 2009; G. Wang, Guangwei, Zenggang, Yihua, & Conghuan, 2013). The objects detected can be used for various application such as video retrieval (Yutaka & Yutaka, 2012), change detection (Niemeyer, Marpu, & Nussbaum, n.d.), segmentation (P. Wang & Ping, 2010), in perception of motion(Hock & Nichols, 2013) and category level detection (Hock & Nichols, 2013; Ponce et al., 2007). Local edge features and region based methods seems to indicate a better quality of edge features (Hock & Nichols, 2013; Ponce et al., 2007; Tang Xusheng et al., 2009). The the edge detection methods on color cue has proved to be a useful tool for object recognition in images (Jiqiang Song, Jiqiang, Min, & Lyu, n.d.; Tsang & Tsang, n.d.). Grouping and fusion of edge features have also a positive impact on the detection of objects (Amit, 2002; Antoniu & Eduard, 1994; Chunxin, Wang, & Xu, 2009; J. Kim & Chen, 2003; Mednieks & Ints, 2008; Meurant, 1992; Mu, Nan, Xin, & Ziheng, 2015; Srikantha, Abhilash, & Juergen, 2014; Stathaki, 2011; Torralba, Antonio, Murphy, & Freeman, 2006).

Object tracking is another important area in object detection research that make use of a wide range of edge detection techniques(Chunxin et al., 2009; Gao et al., n.d.; Wei Jyh Heng & Ngan, 2001; Mukherjee et al., 2010). In tracking problems, the speed of tracking and detection of edges become an important problem, and that leads to development of hardware friendly approaches (W. J. Heng & Ngan, n.d.). Optical flow algorithms have been used along with particle filters to implement real-time tracking(Mae & Shirai, n.d.; Shirai et al., 1996).

# Medical Imaging

Medical images contain several structural and texture details that relate to identifying the health of an individual(Ardeshir Goshtasby & Stavri, 2007; Bourne, 2010; Alex Pappachen James & Dasarathy, 2014; Singh & Khare, 2013). Like object detection, the structural information can be used to segment areas with images created with different modalities: MRI (Laishram, Romesh, Kumar, Anshuman, & Prakash, 2014; Taghizadeh, Moslem, & Mahboobeh, 2011), CT (Leondes, 2003), and ultrasound (Hopp, Zapf, & Ruiter, 2014; Leondes, 2003). Gradient filters such as based on canny operator (Zheng, Zhou, Zhou, & Gong, 2015), and active shape models (Arámbula Cosío, Acosta, & Edgar, 2015) find use in ultrasounds. In radiology images, use of knowledge (Riste-Smith, 1990), curvatures(Leondes, 2003), surface fitting (Hopp et al., 2014), and gradients (Hopp et al., 2014; Zheng et al., 2015) are used in combination with edge detectors for implementing segmentation algorithms.

Multimodal fusion of medical images also utilize several edge feature detection methods to combine and enhance the quality of features. The spatial structural cues (Jia, Huang, & Wang, 2014; Zheng et al., 2015), clustering methods (Ergen, 2014), and image registration (Wein, Röper, & Navab, 2005) are some of applications of edge features applied in image fusion at feature and decision levels.

# Discussions

Edge detectors have a long history of research in neural understanding for translations to realistic computer algorithms. The inspiration of the edge detectors originate from the understanding of primate brain, and is understood to be a very complex process. While there have been several works that try to understand the basic mechanisms of edge formation in human brain aiming to model the spatiotemporal process (Erlikhman & Kellman, 2015; McIvor, 1988); (Cantoni, 2013; Rosenfeld, 2014). In image processing, majority of the edge detection research pursue under the assumption of static conditions (Canny & John, 1987; Davies, 2012; Pau, 1990), and do not have any dependency on the time. In dynamic situations, when the edge changes are spatiotemporal the edge information echoes the cues of early vision, and is closely linked to the ability of the human to identify objects and cluster them that form the basis of intelligent vision systems.

A majority of the real-time implementation of edge detection involves convolution operator in a basic moving window spatial filtering approach ("Gradient Filters," n.d.; Y. Lee & Yunwoo, 2000; Mathews & Xie, n.d.). In such filters, the speed of processing closely depends on the hardware architecture and complexity of the method to improve it against the robustness to the noise. The speedup can be achieved using FPGA(Maheshwari, S.S.S.P., & Poonacha, n.d.; Qi, Haibing, Jianlan, & Song, 2010; Szlachetko, Boguslaw, & Andrzej, 2007) or custom made VLSI architectures(Bayoumi, 2012; Serrano-Gotarredona, 1997; Yuschik, Matthew, & Hideaki, 1985). There also exists a class of memory based approaches that inspire from neural architectures that can implemented with digital memory, flash cells (Alex Pappachen James, Pachentavida, & Sugathan, 2014) or memristors(Y.-J. Liu et al., 2009; Maan, Kumar, Sherin, & James, 2015; Mousse, Cina, & Ezin, 2015; Rajendran, Jeyavijayan, Harika, Ramesh, & Rose, 2012; Weaver, 1975).

Figure 2 provides the histogram of the published works that relate to the development and use of edge detection methods from 1980 till 2015. In the last decade the growth has been significant due to the increase in the computational capabilities and the compounding need to have intelligent vision systems and to gain a deeper understanding of the working of human vision.

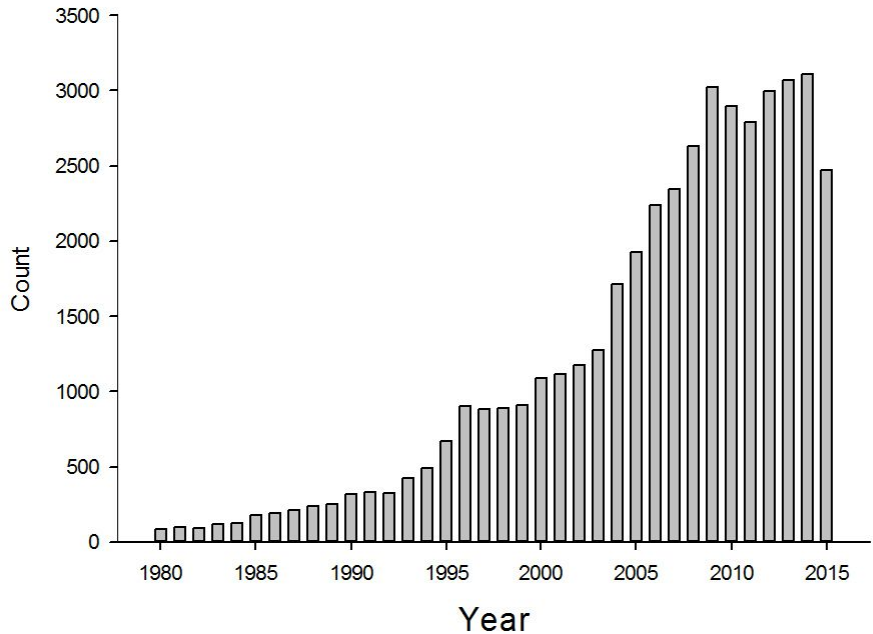

Figure 2: The indicative growth of literature based on the edge detection techniques and its applications. *Source:Scopus*.

# Conclusions

In summary, there exists a wide range of ideas emerging from the foundational studies in perception to neuroscience to that of signal processing, that aim to decode the concepts of edge information in images. The practical implementation of the edge detectors that are immune to variabilities in noise remains a major problem in image recognition. Edges in the images remain to be the most distinct and useful information for pattern recognition, and its encoding in different forms encourage the development of high speed techniques that can be used in real-time applications. The list of applications where edge detectors will be applied in pattern recognition are only set to grow since the camera based automation for monitoring and intelligent processing are on rise supported through the growing maturity of the technological computing and communication infrastructure.